\documentclass[conference]{IEEEtran}
\IEEEoverridecommandlockouts

\usepackage{cite}
\usepackage{amsmath,amssymb,amsfonts}
\usepackage{algorithmic}
\usepackage{graphicx}
\usepackage{textcomp}
\usepackage{xcolor}

\usepackage{hyperref}
\usepackage{tabularx} 
\usepackage{booktabs} 
\usepackage{url}

\def\BibTeX{{\rm B\kern-.05em{\sc i\kern-.025em b}\kern-.08em
    T\kern-.1667em\lower.7ex\hbox{E}\kern-.125emX}}
\begin{document}

\definecolor{darkgreen}{rgb}{0.0, 0.75, 0.0}
\newcommand{\better}[1]{\textcolor{darkgreen}{#1}}
\newcommand{\less}[1]{\textcolor{red}{#1}}

\newcommand{\ar}[1]{\textcolor{blue}{#1}}
\newcommand{\as}[1]{\textcolor{teal}{Ashish: #1}}
\newcommand{\xc}[1]{\textcolor{orange}{\textit{Xiwen: #1}}}
\newcommand{\hw}[1]{\textcolor{violet}{Hao: #1}}
\newcommand{\arr}[1]{\textcolor{red}{\textit{Razi: #1}}}

\title{
FLAME Diffuser: Wildfire Image Synthesis using Mask Guided Diffusion
}

\author{\IEEEauthorblockN{1\textsuperscript{st} Hao Wang, Sayed Pedram Haeri Boroujeni}
\IEEEauthorblockA{\textit{School of Computing} \\
\textit{Clemson University}\\
Clemson, SC, USA \\
{hao9, shaerib}@g.clemson.edu}
\and
\IEEEauthorblockN{2\textsuperscript{nd} Xiwen Chen}
\IEEEauthorblockA{\textit{School of Computing} \\
\textit{Clemson University}\\
Clemson, SC, USA \\
xiwenc@g.clemson.edu}
\and
\IEEEauthorblockN{3\textsuperscript{rd} Ashish Bastola}
\IEEEauthorblockA{\textit{School of Computing}\\
\textit{Clemson University}\\
Clemson, SC, USA \\
abastol@g.clemson.edu}
\and
\IEEEauthorblockN{4\textsuperscript{th} Huayu Li}
\IEEEauthorblockA{\textit{Electrical and Computer Engineering} \\
\textit{The University of Arizona}\\
Tucson, AZ, USA \\
hl459@arizona.edu}
\and
\IEEEauthorblockN{5\textsuperscript{th} Wenhui Zhu}
\IEEEauthorblockA{\textit{School of Computing and Augmented Intelligence} \\
\textit{Arizona State University}\\
Tempe, AZ, USA \\
wzhu59@asu.edu}
\and
\IEEEauthorblockN{6\textsuperscript{th} Abolfazl Razi}
\IEEEauthorblockA{\textit{School of Computing} \\
\textit{Clemson University}\\
Clemson, SC, USA \\
arazi@clemson.edu}

}

\maketitle

\begin{abstract}
Wildfires are a significant threat to ecosystems and human infrastructure, leading to widespread destruction and environmental degradation. Recent advancements in deep learning and generative models have enabled new methods for wildfire detection and monitoring. However, the scarcity of annotated wildfire images limits the development of robust models for these tasks. In this work, we present the \textbf{FLAME Diffuser}, a training-free, diffusion-based framework designed to generate realistic wildfire images with paired ground truth. Our framework uses augmented masks, sampled from real wildfire data, and applies Perlin noise to guide the generation of realistic flames. By controlling the placement of these elements within the image, we ensure precise integration while maintaining the original image’s style. We evaluate the generated images using normalized Fréchet Inception Distance (nFID), CLIP Score, and a custom CLIP Confidence metric, demonstrating the high quality and realism of the synthesized wildfire images. Specifically, the fusion of Perlin noise in this work significantly improved the quality of synthesized images.
The proposed method is particularly valuable for enhancing datasets used in downstream tasks such as wildfire detection and monitoring.

The resources for accessing the code and datasets are provided at the link: 
\url{https://arazi2.github.io/aisends.github.io/project/flame}

\end{abstract}

\begin{IEEEkeywords}
Image Synthesis, Image Processing, Diffusion Models, Wildfire Imaging, Forestry
\end{IEEEkeywords}

\section{Introduction}
\label{sec:intro}

Over the past decades, wildfires have caused devastating damage to natural ecosystems, human settlements, and infrastructure \cite{boroujeni2024comprehensive}. Wildfires significantly reduce air and soil quality, disrupt biodiversity, and increase the risk of secondary disasters such as flooding and landslides. Moreover, they accelerate climate change by releasing greenhouse gases into the atmosphere through the destruction of vegetation \cite{shakesby2006wildfire,ghali2022deep,mohapatra2022early,chen2022wildland}. Given the severity of these consequences, wildfire detection and management have become critical areas of research. However, timely detection and effective response remain major challenges. Traditional methods, which rely on satellite-based remote sensing and UAVs, have proven effective for monitoring and tracking wildfires. While these technologies, including thermal imaging and image-based recognition, are widely used in real-time wildfire detection, their ability to meet the demand for immediate action is still limited by operational constraints and deployment time \cite{casas2023assessing,boroujeni2024ic}.

Recent advances in Deep Learning (DL), particularly in object detection and image classification, have significantly enhanced wildfire detection accuracy \cite{diwan2023object,wang2022real,liu2020uav}. However, due to the rarity and dangerous nature of wildfires, acquiring sufficient training data remains a major bottleneck \cite{ghali2022deep, toulouse2017computer}. The scarcity of annotated wildfire images limits the development of robust models for downstream tasks, such as improving object recognition precision or classification accuracy in wildfire scenarios \cite{zhu2020deformable, chen2022wildland, diwan2023object}.
In response to these challenges, generative models, especially diffusion models \cite{rombach2022high}, have emerged as powerful tools for data augmentation \cite{nguyen2024dataset}. These models offer significant improvements over traditional methods such as Generative Adversarial Networks (GANs) by providing better image diversity, realism, and control over the generated outputs \cite{loey2020deep,zhu2023otre,boroujeni2024ic}. Diffusion models are highly versatile, enabling the generation of random high-quality images, images conditioned on text prompts, and even images guided by both input images and textual descriptions. This flexibility makes diffusion models particularly well-suited for creative applications and data augmentation across various fields \cite{meng2021sdedit,zhang2023adding,zhou2023denoising,xu2023diffscene,zhang2024transparent}.

\begin{figure}[ht]
    \centering
    \centerline{\includegraphics[width=1\linewidth]{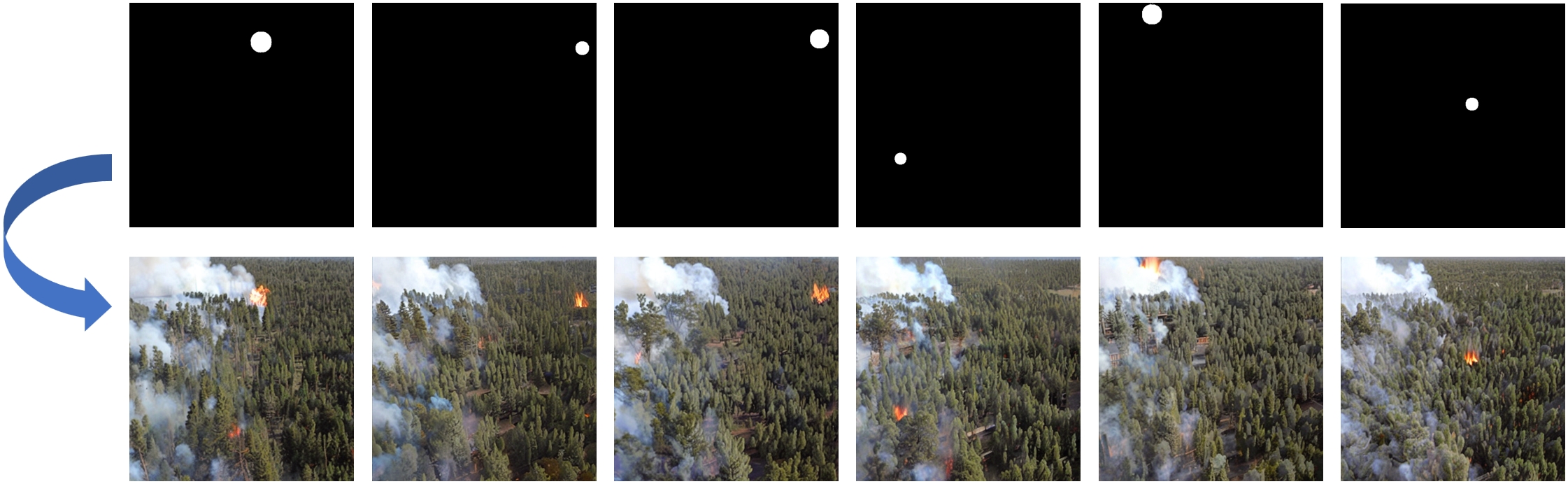}}
    \caption{The concept of wildfire image synthesis, where the masks can guide the location of flames in the generated images.}
    \label{fig:concept}
\end{figure}

Despite their potential, generating specific content, such as wildfire elements in an image, remains a significant challenge in diffusion models \cite{zhang2023adding}. Unlike text-to-image generation, image-to-image synthesis involves more complex dependencies, as the output is highly influenced by both the input image and the text prompt \cite{rombach2022high}. In wildfire image generation, we aim to synthesize images that not only retain the natural style of the scene but also incorporate precise information on wildfire locations. 
As illustrated in Figure \ref{fig:concept}, our approach to wildfire image synthesis leverages image-to-image diffusion to produce realistic images that maintain the style of the original input while using a mask to guide the generation of elements (e.g., flames) \cite{bashkirova2023masksketch}. 
While diffusion models such as Stable Diffusion’s inpainting feature are capable of generating new content in masked areas, it is revealed that there exists a limitation in controlling the exact placement and appearance when the elements and existing background may not become coherent \cite{liu2024structure}.
Recently, ControlNet has been proposed to address some of these challenges by providing more granular control over the structure and details of generated images \cite{zhang2023adding}. However, even with ControlNet, achieving precise control over wildfire elements, such as flames, proved difficult \cite{meng2021sdedit,couairon2022diffedit}. In our later experiments, although the generated images included fire, determining the exact location of the wildfire within the image remained unresolved.

To address these limitations, we propose the \textbf{FLAME Diffuser}, a simple yet effective solution by introducing a specialized mask into the initial input image, which can guide the diffusion process to gradually transform the masked area into wildfire elements. We discovered that, in image-to-image tasks, the style and content of the output image are largely determined at the beginning of the generation process \cite{everaert2023diffusion,everaert2024exploiting}. 
Therefore, our approach leverages this insight by manipulating the initial RGB image in pixel space, ensuring that the initial input already resembles wildfire-ready features. The subsequent denoising and text-guided steps further refine the image, producing both realistic and diverse wildfire images. 
This method allows us to generate images with high diversity while simultaneously providing precise information about wildfire locations. More importantly, the proposed framework is train-free and requires no further tuning of any modes, which is highly practical for downstream tasks such as wildfire detection and monitoring.

In summary, the major contributions of this paper are summarized as follows:

\begin{itemize}
\item We present the \textbf{FLAME Diffuser}, a training-free diffusion-based framework for generating ground truth paired wildfire images. This framework utilizes style images and augmented masks to provide precise control over the placement and integration of wildfire elements (e.g., flames) in the existing image.
    \item  We propose a mask augmentation method capable of generating masks with varied complexity, geometry, and textures, enhancing the diversity and realism of the synthesized wildfire elements.  
    \item We present an automated annotation method using CLIP, demonstrating how \textbf{CLIP Confidence} can be applied to evaluate the success and accuracy of image synthesis.
\end{itemize}

The remaining paper is organized as follows: Section \ref{sec:RelatedWork} provides a comprehensive review of related work in the field of wildfire datasets, diffusion models, and text-image architecture. In Section \ref{sec:Methodology}, we present the details of the methodology and implementation of our proposed approach. The experimental results are discussed in Section \ref{sec:Experiment}.

\section{Related Work}
\label{sec:RelatedWork}

In this section, we provide a comprehensive review of recent advancements in wildfire detection and image analysis, which are crucial for understanding the context and relevance of our proposed approach. We focus on three key areas: wildfire datasets, diffusion models, and image-text architectures. By examining these areas, we aim to highlight both the challenges and the progress made in related fields, offering insights into how our approach addresses these issues.

\subsection{Wildfire Datasets}
There exist a few publicly available datasets that provide aerial imagery during an active forest fire. We investigated such datasets and found FLAME1\cite{shamsoshoara2021aerial} and FLAME2 \cite{afghah2022,chen2021wildland} most appropriate for this study since they provide a comprehensive collection of fire aerial images in both RGB and IR domains. 

\begin{figure}[htbp]
    \centering
    \centerline{\includegraphics[width=1\columnwidth]{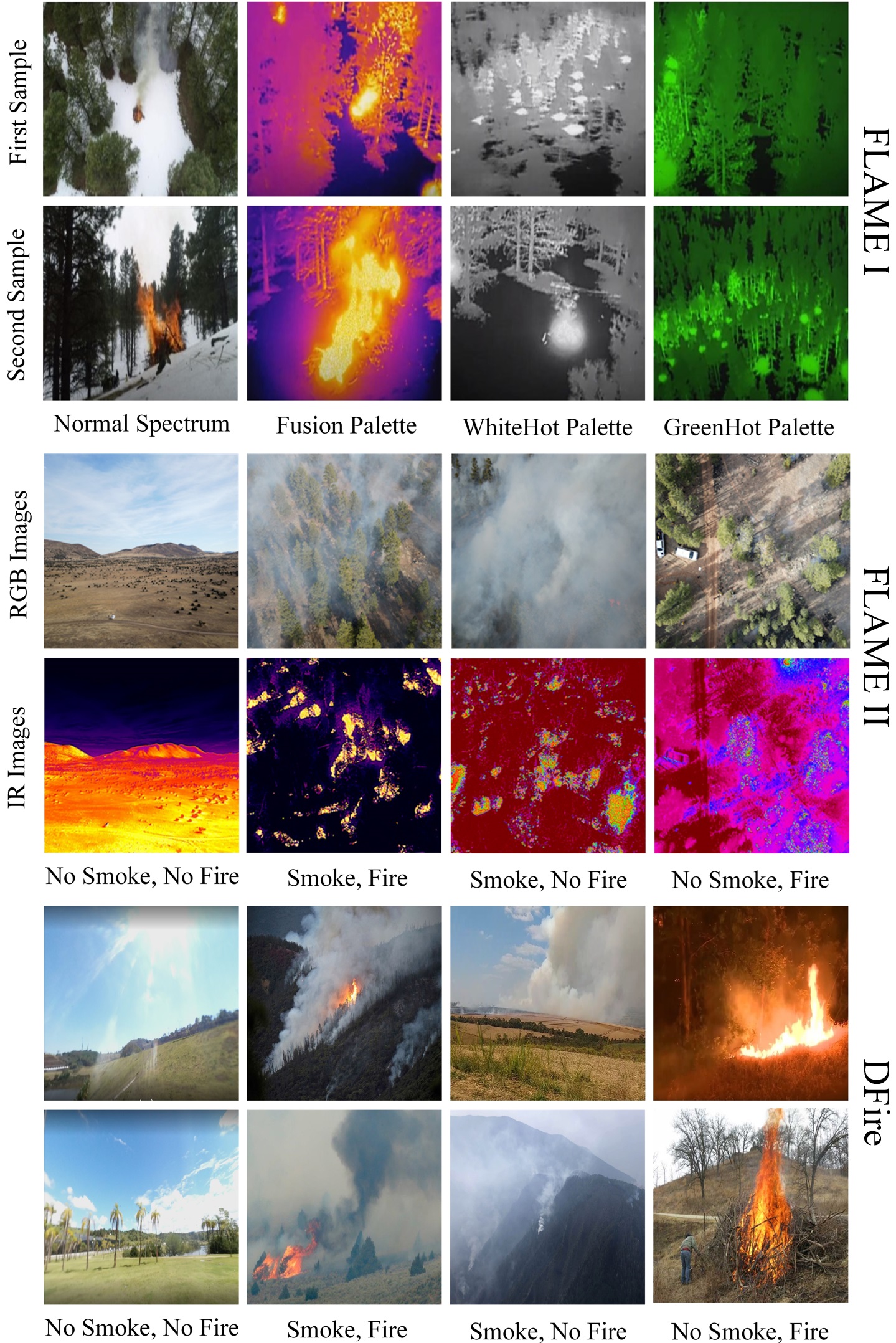}}
    \caption{Sample images of the FLAME1, FLAME2, and DFire datasets.}
    \label{fig:dataset}
\end{figure}

The FLAME1 dataset provides multiple aerial video recordings collected by drones throughout a prescribed burn operation. Each video has been transformed into individual image frames to enable image-based analysis according to the Frames Per Second (FPS) rate. The dataset includes four different types of visual representations, including normal spectrum, fusion, white-hot, and green-hot schemes. Figure \ref{fig:dataset} presents some samples of the FLAME1 dataset to provide a better insight into the visual contents of the collected fire images.


Likewise, the FLAME2 dataset includes video recordings and thermal heatmaps captured by infrared cameras \cite{chen2022wildland}. The IR and RGB modes are fully synchronized, enabling the IR images to serve as a ground truth representation. The captured videos and images are manually annotated by three experts, and labeled frame-wise to provide a comprehensive dataset for various image-based tasks. This dataset includes 53,451 pairs of $254p\times254p$ RGB/IR images labeled with three classes: (1) Flame with Smoke, (2) Flame with No Smoke, and (3) No Flame with No Smoke. Figure \ref{fig:dataset} presents some samples of dual RGB/IR images from the FLAME2 dataset to provide a better insight into the dataset's characteristics. However, we note that these datasets are often collected in limited scenarios. For example, FLAME2 recorded the prescribed burns and wildfires in roughly only three or four areas. As mentioned in Section I, DL models learned on these datasets may lack generalizability to other scenarios with different environments.

The D-Fire dataset \cite{de2022automatic} is a comprehensive image dataset designed specifically for machine learning and object detection algorithms in the context of fire and smoke detection. It contains over 21,000 images, categorized into four distinct groups: 1,164 images featuring only fire, 5,867 images with only smoke, 4,658 images showing both fire and smoke, and 9,838 images with neither, serving as a control group. The dataset includes a total of 14,692 bounding boxes for fire and 11,865 for smoke, all annotated in the YOLO format with normalized coordinates. 
This extensive and well-annotated dataset is particularly valuable for training and evaluating models in fire and smoke detection, making it a significant resource for research and development in this critical area of computer vision and emergency response technology. Figure \ref{fig:dataset} presents some samples of various images from the D-Fire dataset to provide a better insight into the dataset's characteristics.


\subsection{Latent Diffusion}

Latent diffusion represents an innovative technique in the realm of generative models designed to synthesize novel data samples. These models have been applied in diverse applications, particularly in image synthesis, providing better quality and easier content control \cite{rombach2022high}. They work by iteratively refining a noisy image until a clear, high-quality image is produced, allowing for the integration of detailed features and fine styles \cite{guo2024make}. This iterative process makes diffusion models particularly suitable for generating realistic images. The foundational principle of latent diffusion models is inspired by the diffusion process observed in physics \cite{ho2020denoising}. Unlike conventional diffusion models that manipulate data directly in the pixel space, latent diffusion models operate within a condensed and abstract representation of the data, known as the latent space. This approach enhances the efficiency of the model and contributes significantly to generating diverse outputs \cite{rombach2022high}.
A classical latent diffusion model consists of three key components: the \textbf{Variational Autoencoder (VAE)}, which compresses high-dimensional input images into a lower-dimensional latent space to reduce computational complexity; the \textbf{denoising U-Net}, which refines noisy latent representations into coherent outputs; and an \textbf{image-text tokenizer} that bridges the gap between textual prompts and images \cite{radford2021learning}. 

Despite these advantages, diffusion-based image synthesis faces challenges in precisely controlling specific elements within the images. Additionally, domain shift issues in image-to-image synthesis can lead to inconsistencies in the generated datasets \cite{huang2023domain, lauenburg20233d}.

\subsection{Image-Text Architecture}

CLIP (Contrastive Language–Image Pre-training) is a cutting-edge model developed by OpenAI that plays a crucial role in the advancement of text-image connections \cite{radford2021learning}. This architecture bridges the gap between textual descriptions and visual content, enabling machines to understand images in the context of natural language descriptions. CLIP is designed to learn visual concepts from natural language descriptions by training on a wide variety of images and their corresponding text captions, allowing it to understand and generate content across different modalities. At the heart of CLIP's effectiveness is contrastive learning, where the model learns to recognize which images correspond to which text descriptions among a batch of incorrect pairings, thus creating a rich representation of images and text. In other words, the emergence of contrastive learning significantly boosts multimodal representations, fostering rapid advancements in text-image understanding. 

One of the key strengths of CLIP is its ability to generalize across a broad range of tasks without task-specific training data. This versatility means that CLIP can be applied to numerous applications, from generating images based on textual descriptions to improving search engines by understanding the content of images in relation to text queries. CLIP's capabilities have been instrumental in developing advanced text-to-image models such as DALL·E \cite{ramesh2021zero}, which can generate highly detailed and creative images from textual descriptions.

\begin{figure*}[htpb]
    \centering
    \centerline{\includegraphics[width=1\linewidth]{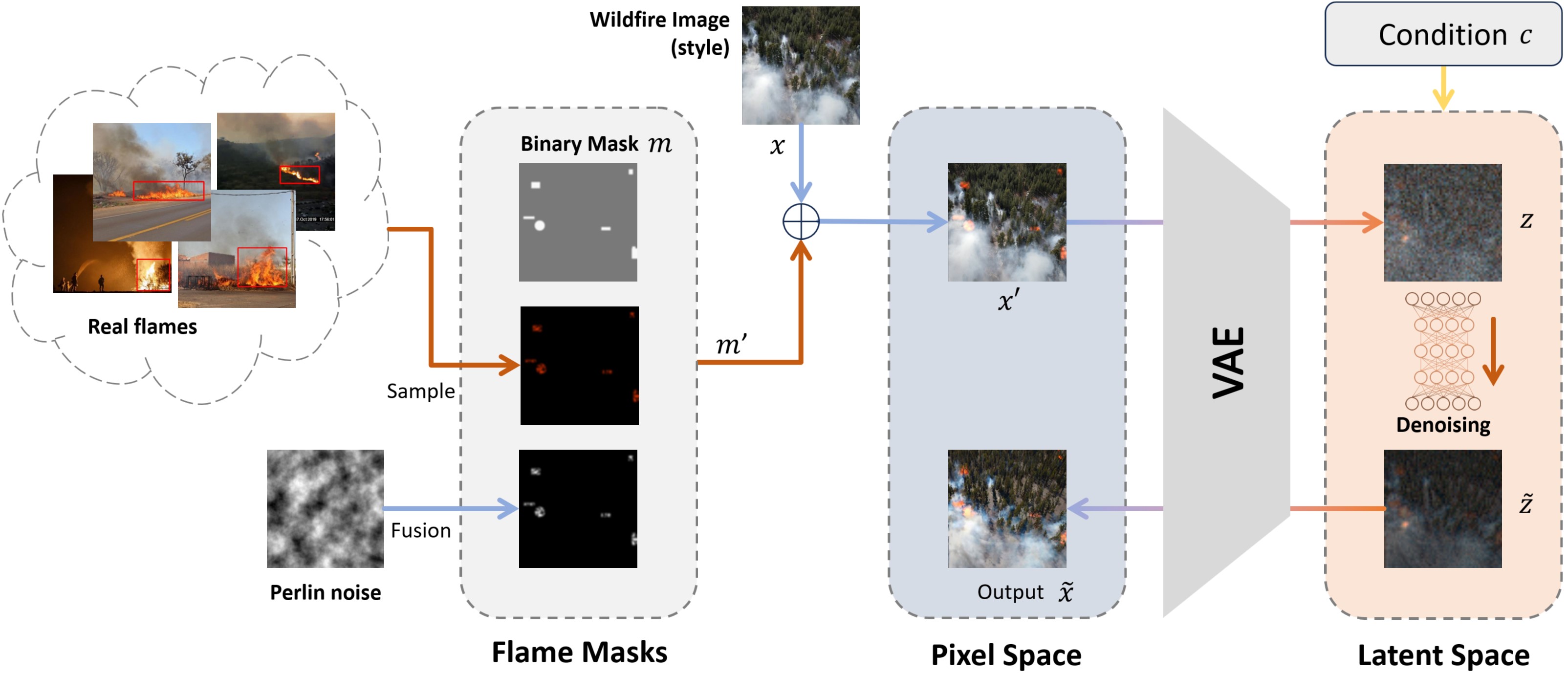}}
    \caption{Framework of the FLAME Diffuser.}
    \label{fig:diffuser}
\end{figure*}

\section{Problem Formulation}
\label{sec:prob}
In this section, we discuss the essential challenges in generating wildfire elements using diffusion models in image-to-image (I2I) systems and our approach to addressing them. Specifically, we highlight the difficulty of maintaining original image styles while introducing precise control over specific elements, such as wildfires, in the generated image.

\subsection{Limitations of Existing Image-to-Image Diffusion}
\label{sec:limit}
\begin{figure}[ht]
    \centering
    \centerline{\includegraphics[width=0.8\columnwidth]{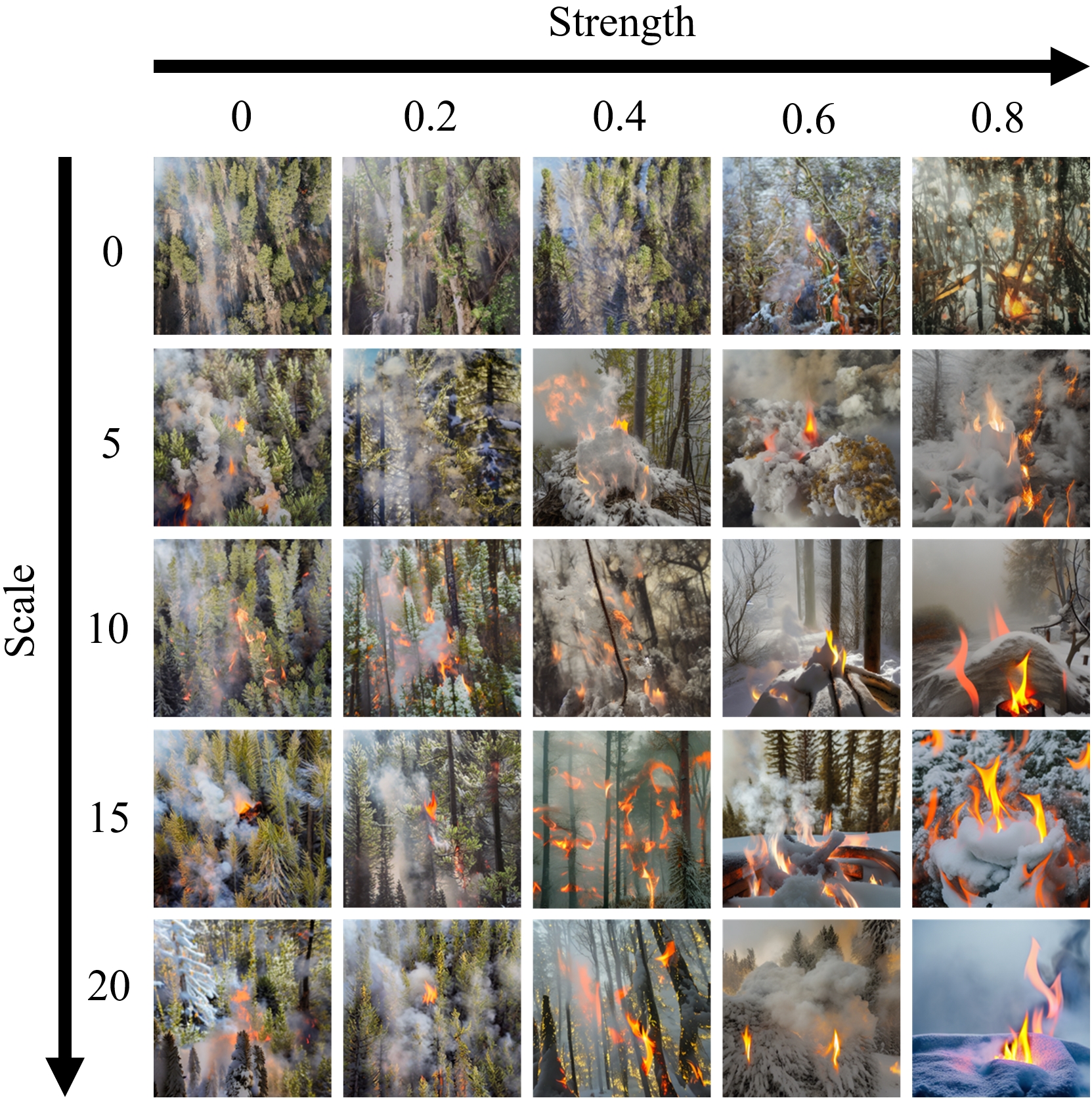}}
    \caption{Wildfire image synthesis in terms of text prompt scale and image denoise strength. Text prompt: \textit{'Wildfires in the snow, flame in the smoke, high-resolution, photo realistic.'}}
    \label{fig:factor}
\end{figure}

Conventional image-to-image generation techniques face significant challenges when tasked with adding specific content, such as wildfires, at predefined locations within an image, while preserving the original style. This issue becomes more pronounced as the requirements for content control increase. Our initial experiments, involving various parameter combinations and settings, demonstrate that stronger denoising and text guidance are necessary to introduce specific elements into the generated image. However, these adjustments often come at a cost: increased denoising strength can result in a loss of image realism, while excessive reliance on text prompts can cause stylistic inconsistencies, leading to visual artifacts and unnatural transitions between the inserted elements and the original image, as shown in Figure \ref{fig:factor}.

For instance, in wildfire detection, it is crucial that the generated images contain not only visually accurate fire elements but also retain the environmental context, such as forests or landscapes, without noticeable stylistic disruption. Yet, in current diffusion-based models, the stronger the focus on generating the wildfire elements, the more the background realism suffers. This trade-off highlights a fundamental limitation in existing systems: it is difficult to balance fine-grained content control with the preservation of overall image fidelity \cite{wu2023diffumask}.

\subsection{Image Manipulation}
\label{sec:manipulation}
Previous studies have shown that the final image's content and style are closely tied to the initial image used in the generation process \cite{everaert2023diffusion}. Many state-of-the-art methods attempt to address content control by manipulating the latent tensor in the latent space, which is typically derived from a Variational Autoencoder (VAE) \cite{rombach2022high}. While this approach has the potential to influence the generation process more subtly, direct manipulation of these latent tensors is risky due to their complex, highly compressed, and trained nature. Operations performed in the latent space often lead to image degradation or collapse, particularly when dealing with highly trained VAE networks that encode information in an abstracted and compressed form. Ensuring safe manipulation of these tensors typically requires additional training or fine-tuning of the model, which can be resource-intensive and difficult to implement in practice \cite{ho2020denoising,everaert2023diffusion}.

Given these challenges, our proposed method avoids latent space manipulation altogether. Instead, we opt to guide the generation process through pixel-level operations in the input space. This decision reduces the risk of image degradation but introduces its own set of challenges, including the potential for abrupt pixel transitions that can compromise the realism of the generated image.



\section{Methodology}
\label{sec:Methodology}


The architecture of our proposed method is illustrated in Figure \ref{fig:diffuser}. First, RGB values representing real flames are sampled from the DFire dataset, and then blended into randomly generated binary masks. Perlin noise is then applied and fused with these blended masks to introduce smoother variations. The original wildfire images from the FLAME1 and FLAME2 datasets are fused with the augmented masks to create the input for the generative process. This fused image is encoded by a pre-trained Variational Autoencoder (VAE), and a pre-trained diffusion model (SDv1.5) is used to denoise the latent tensor. During the denoising process, a text prompt (serving as a condition) guides the synthesis of the image. The denoised tensor is then decoded back into an image by the same VAE, producing both a wildfire image and its corresponding ground-truth mask.

The process can be further torn down into several key steps:




\subsection{Mask Generation}
\label{sec:noise}
To define the regions where fire elements should appear in the image, we use mathematical algorithms to generate random masks. These masks can then be enhanced by applying various image processing techniques such as color transformation, noise addition, and domain warping to make them more dynamic and diverse. This ensures that the synthesized images exhibit variability in flame appearance and placement.

Specifically, we implemented different types of masks for fusion: \textbf{binary} masks, \textbf{colored} masks, \textbf{noise} masks, and domain-warping masks or \textbf{Perlin} masks. 
The binary masks are simple geometrical shapes (e.g., rectangles, circles) generated mathematically. 
To generate colored masks, we sample the RGB value distribution of real fires from the DFire dataset. As shown in Figure \ref{fig:diffuser}, the sampled RGB value is then used to replace the color tone of the binary mask $m$.
The noise masks are created by adding Gaussian noise to the colored masks, introducing additional random variations. 
Lastly, the domain warping masks are generated by fusing the colored masks with a special type of noise $-$ Perlin noise, which produces smoother and more natural variations in the mask.
Perlin noise, a type of domain warping noise, was originally developed for computer graphics and is a gradient noise function known for generating smooth, natural-looking textures \cite{green2005implementing}. Unlike random noise such as Gaussian noise, Perlin noise creates coherent structures with gradual transitions, making it ideal for simulating organic patterns like clouds, terrain, and fire. 


\begin{figure*}[ht]
    \centering
    \centerline{\includegraphics[width=1\linewidth]{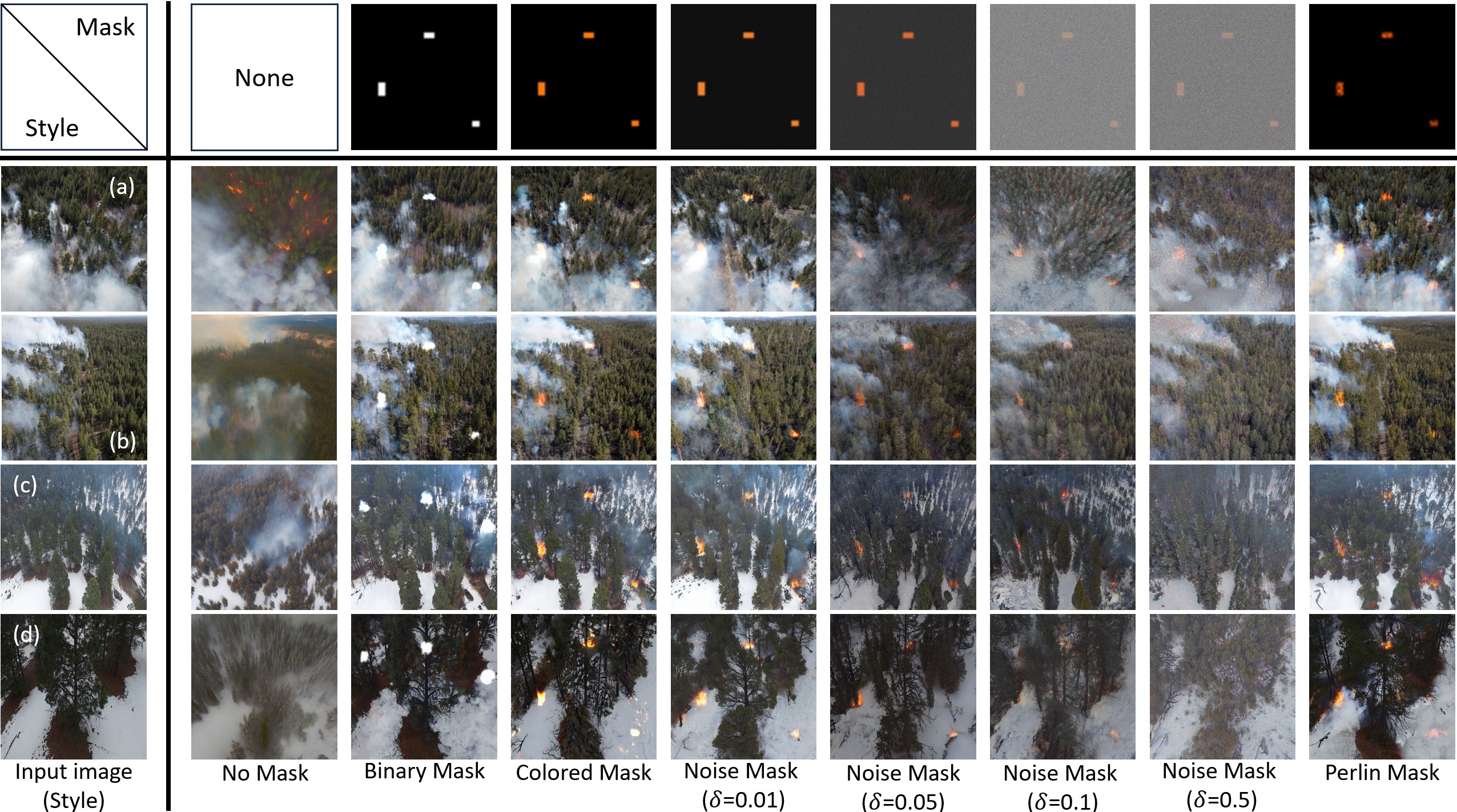}}
    \caption{Sample images from FLAME diffuser. The top row is the input mask, and left leftmost column is the input style image. (a) and (b) are two different scenes from the FLAME2 dataset, (c) and (d) are another two scenes from the FLAME1 dataset.}
    \label{fig:grid}
\end{figure*}

\subsection{Mask-Image Diffusion}
As shown in Figure \ref{fig:diffuser}, an image $x$ is sampled from the FLAME1 and FLAME2 datasets. The augmented mask $m'$ is then fused with the sampled real image $x$, forming a new composite input image $x'$, which serves as the input for the subsequent stages of the diffusion process:
   \[
   x' = x \oplus m'
   \]
   where $ \oplus$ denotes element-wise addition between the sampled image $x$ and the augmented mask $m'$.

The composite image $x'$ is passed through a Variational Autoencoder (VAE) encoder $\mathcal{E}$, which transforms the image into a latent representation $z$. The noised mask $m'$ serves as a control matrix, containing context-specific information that will guide the subsequent denoising process:
   \[
   z = \mathcal{E}(x')
   \]

The latent variable $z$ is then passed through a denoising U-Net, which performs multiple steps of denoising to gradually refine the image. During this process, the mask $m'$ and a well-constructed text prompt $c$ are used to guide the denoising process, ensuring that the synthesized fire elements appear in the desired regions of the image while maintaining realism and consistency with the surrounding environment. The U-Net produces a new latent variable $\tilde{z}$ after n steps of denoising, and the refined latent variable $\tilde{z}$ is decoded back into an image using the VAE decoder $\mathcal{D}$, producing the synthesized image $\tilde{x}$:
   \[
   \tilde{x} = \mathcal{D}(\tilde{z}) = \mathcal{D}(z; m', c)
   \]
   The mask $m'$ and the text prompt $c$ play crucial roles in shaping the final image, ensuring that the fire elements blend seamlessly with the real image background while preserving the original style.


Key to this process is the transformation applied to the mask $m$ before it is fused with the image. The noise added to the mask serves to smooth transitions between the mask and the real image, while the use of domain warping techniques ensures that the generated fire elements appear natural and realistic in context.
Our method manipulates the images in the pixel space directly, making the initial composite image $x'$ more closely resemble the desired wildfire structure. This approach provides better control over the content and appearance of the final output image without requiring fine-tuning or retraining of the diffusion model.

\section{Experiments}
\label{sec:Experiment}


In this section, we evaluate the quality of the synthesized wildfire images and assess the impact of different mask types. Since the output can be influenced by various parameters during the denoising process, we adjust those unrelated to our contributions to optimal levels. This ensures consistency and prevents stylistic shifts that might disrupt image coherence.

Specifically, we fix the \textbf{denoise strength} at $0.5$ and the text prompt \textbf{guidance scale} at $5$. The following text prompt is used consistently throughout the experiments: \textit{"wildfire with flame and smoke, drone view, photo realistic, high resolution, 4k, HD."} By maintaining these parameters, we isolate the effects of mask types and focus on their influence on the generated images.

\subsection{Qualitative Evaluation}


To ensure a fair comparison, we use images generated with maximum denoise strength ($0.99$) as the \textbf{baseline} to show the maximum performance of the pre-trained diffusion model (SDv1.5). In this baseline setting, mask input is disabled to prevent any interruptions to the image context.

\begin{figure*}[htpb]
    \centering
    \centerline{\includegraphics[width=1\linewidth]{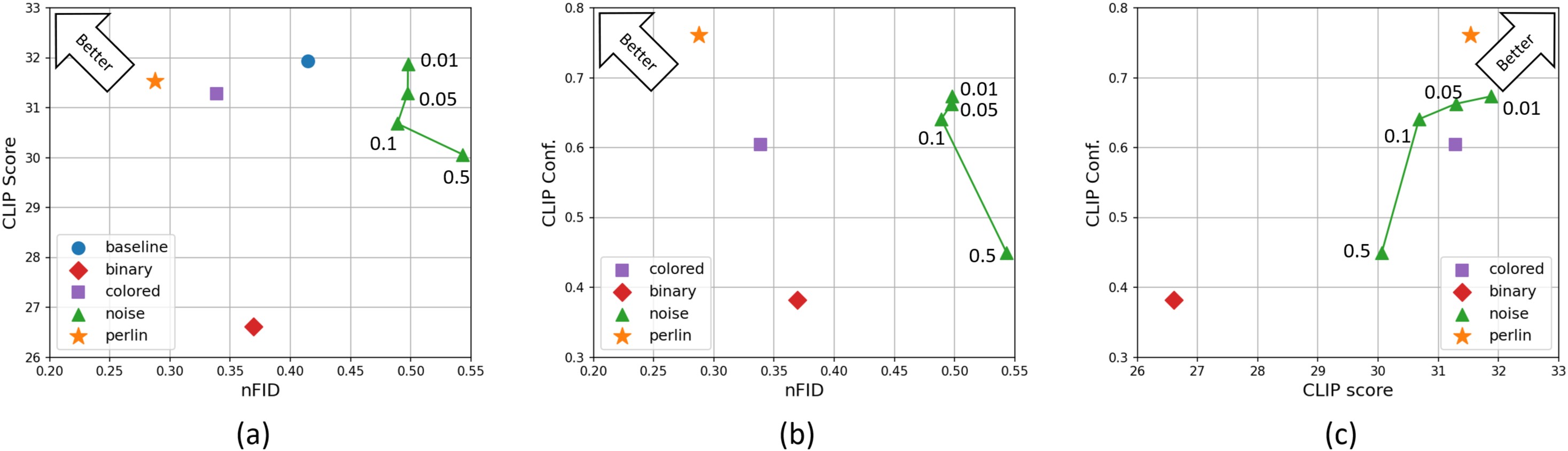}}
    \caption{Joint analysis of nFID, CLIP Score, and CLIP Confidence. (a) is the joint comparison between nFID and CLIP Score, (b) is the joint comparison between nFID and CLIP Confidence, (a) is the joint comparison between CLIP Score and CLIP Confidence.}
    \label{fig:fid}
\end{figure*}

As illustrated in Figure \ref{fig:fid}, the baseline successfully synthesizes flames within the image, but their placement is random due to the fully randomized initial latent tensor. By contrast, with mask-guided image diffusion, the flame locations are well-preserved in the final images, thanks to the controlled guidance provided by the masks.

Moreover, the flames generated with Perlin and noise masks are seamlessly blended into the environment, resulting in a more natural and coherent integration of the wildfire elements into the background compared to the baseline setting.

\begin{figure}[ht]
    \centering
    \centerline{\includegraphics[width=1\linewidth]{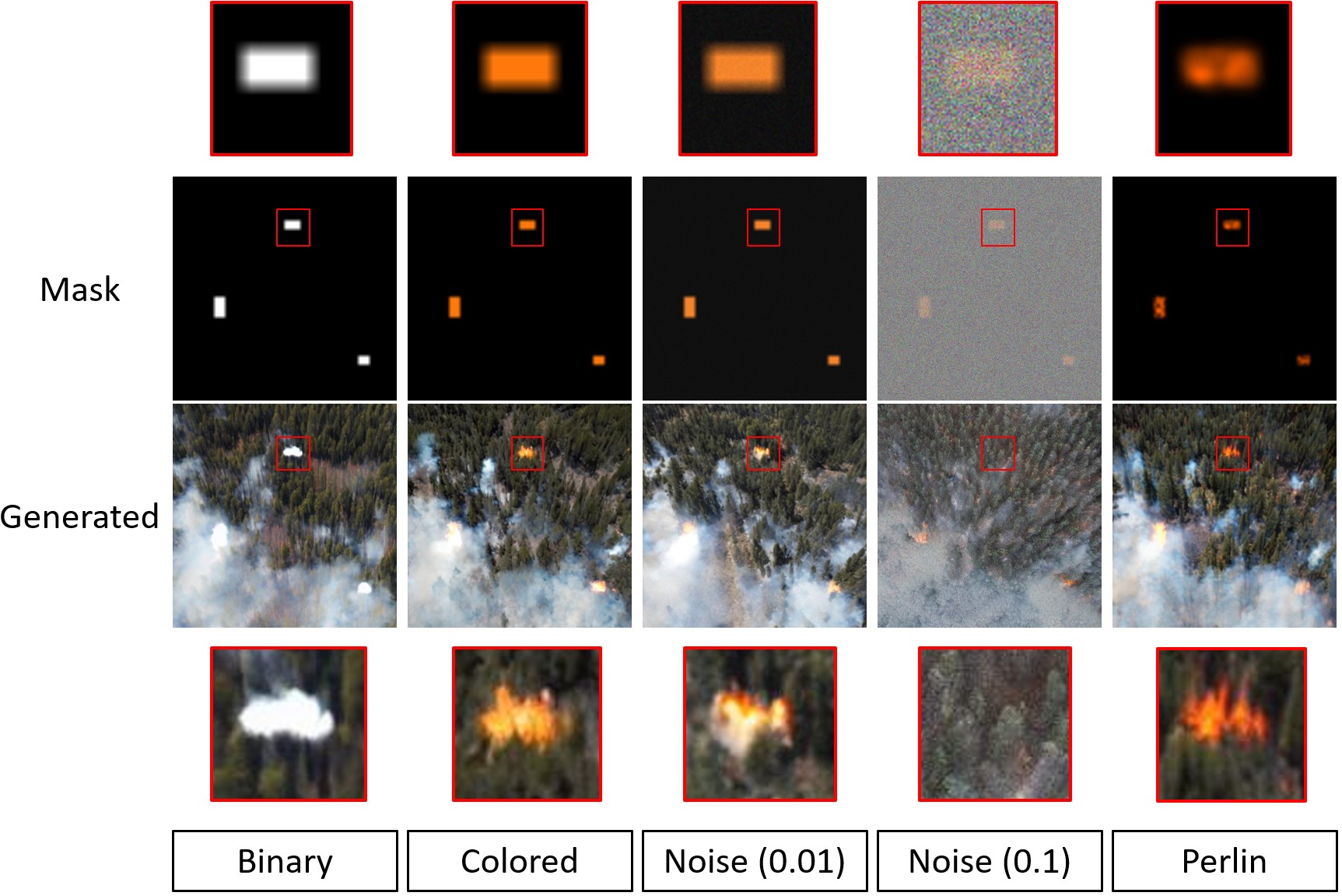}}
    \caption{Wildfire image synthesis with augmented masks.}
    \label{fig:mask}
\end{figure}

Additionally, Figure \ref{fig:mask} demonstrates the impact of different mask types on the synthesized images. For example, binary masks tend to generate smoke or cloud-like elements rather than flames. This is because the input image is fused with the binary mask, and the simplistic white shapes in these masks are difficult to refine into realistic wildfire flames during the denoising process, as discussed in Section \ref{sec:manipulation}. 
In contrast, other mask types (colored, noise, and Perlin) successfully generate realistic flames after denoising. However, it is important to note that increasing the noise ratio can lead to the disappearance of certain elements due to information loss during the denoising process. 


As discussed in Section \ref{sec:manipulation}, while pixel space manipulation offers a safer alternative to latent space operations, it comes with its own challenges. One of the primary concerns is that abrupt pixel changes, introduced by manually adding elements like wildfires, can appear unnatural or inconsistent with the surrounding environment during the diffusion process. Furthermore, the initial size and texture of the manually added elements are crucial in determining the final image quality. If these elements are either too coarse or too fine in detail, the diffusion model may struggle to integrate them smoothly, resulting in visual artifacts or inconsistencies in scale and texture.




\begin{figure}[ht]
    \centering
    \centerline{\includegraphics[width=1\linewidth]{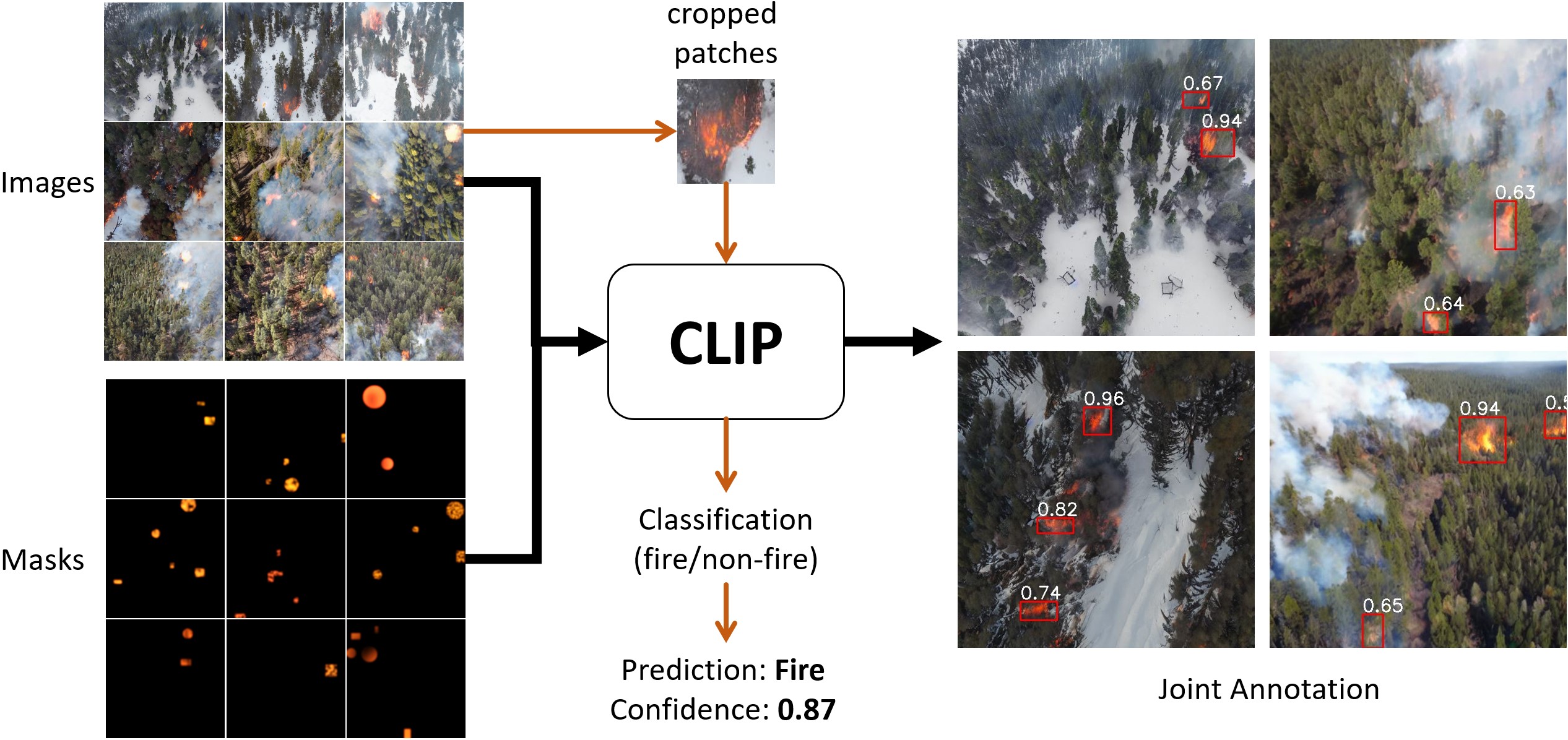}}
    \caption{Using CLIP for classification and confidence identification. An image patch will be cropped according to its mask and will be sent to CLIP for classification, the result will be displayed on the generated image (joint annotation).}
    \label{fig:clip}
\end{figure}

\subsection{Quantitative Evaluation}
We employ three metrics—normalized FID, CLIP score, and CLIP confidence—to evaluate the quality of the generated images and the effectiveness of our proposed method. 

Specifically, \textbf{FID} (Fréchet Inception Distance) measures the similarity between the distributions of real and generated images by comparing their feature embeddings from a pre-trained network \cite{szegedy2016rethinking, wright2022artfid}. Lower FID scores indicate closer alignment to real images, making it a reliable metric for assessing image quality. 
For each setting, we generate $10,000$ images and measure the FID with a total of $11,894$ real images from FLAME1 and FLAME2. We then normalize the FID scores to emphasize performance differences across methods and highlight the effectiveness of our approach \cite{everaert2023diffusion}. 

Meanwhile, the \textbf{CLIP Score} measures the alignment between the generated image and the associated text prompt \cite{radford2021learning}. By computing the cosine similarity between the image and text embeddings, a higher CLIP score indicates that the generated content closely matches the desired textual description, ensuring the generated images align semantically with the prompts. For each setting, we measure the CLIP Score for each image and use the mean CLIP Score of these images to represent the average quality and diversity level of this image subset.

In addition to the CLIP score, we introduce a custom metric—\textbf{CLIP confidence}—designed to measure how accurately the generated elements resemble fire \cite{lin2023clip}. This metric helps determine whether the newly added elements in the images have been successfully synthesized as flames.
We set up a binary classification system using CLIP, categorizing elements as either "fire" or "non-fire." The non-zero areas of the masks are cropped from the generated images and classified by CLIP, as shown in Figure \ref{fig:clip}. If the cropped area obviously contains fire, the CLIP confidence score will be high, reflecting strong alignment between the expected flame and the generated content. 
This metric allows us to evaluate how confidently the model identifies the synthesized elements as flames, providing an additional measure of flame quality. For each generated image, we calculate the average CLIP confidence across all elements to represent the overall flame quality for that subset.

Figure \ref{fig:fid} presents the image quality assessment of the proposed methods. Specifically, Figure \ref{fig:fid}(a) illustrates the joint analysis of normalized FID (nFID) and CLIP Score. The baseline achieves the highest CLIP Score because it generates fully random images, allowing maximum alignment with the text prompt. Although the noise mask can reach a similar CLIP Score to the baseline, the introduction of Gaussian noise reduces both image diversity and quality, with more noise leading to further degradation in context and overall quality. In contrast, the Perlin mask achieves the best balance between image quality, diversity, and context preservation.
Moreover, as shown in Figure \ref{fig:fid}(b), the Perlin mask method also attains the highest CLIP Confidence, indicating that the flames generated with Perlin masks are of superior quality and appear more realistic compared to other mask types. On the other hand, Figure \ref{fig:fid}(c) shows that Gaussian noise negatively impacts both image quality and the coherence of the generated elements.

\begin{table}[htbp]  
    \centering      
    \caption{Image Quality Comparison}
    \resizebox{1\linewidth}{!}{
    \begin{tabular}{lccc} 
        \toprule      
        &nFID $\downarrow$& CLIP Score $\uparrow$& CLIP Conf. $\uparrow$\\\toprule
 Random&  0.41& \textbf{31.93}& NA\\
 Binary Mask
&  0.37& 26.61& 0.38\\
 Colored Mask& 0.33& 31.28&0.60\\
 Noise Mask ($\sigma=0.01$)& 0.50& 31.87& 0.67\\
 Noise Mask ($\sigma=0.05$)& 0.49& 31.29&0.66\\
 Noise Mask ($\sigma=0.1$)& 0.48& 30.68&0.64\\
 Noise Mask ($\sigma=0.5$)& 0.54& 30.06&0.45\\
 Perlin Mask& \textbf{0.28}& 31.53& \textbf{0.76}\\\bottomrule
    \end{tabular}}
    \label{tab:fid}
\end{table}


In summary, as shown in Table \ref{tab:fid}, the proposed Perlin mask method achieved the best image quality ($\textbf{nFID = 0.28}$) without compromising the image-text alignment ($\textbf{CLIP Score = 31.53}$). Its ability to balance style and context makes it superior to other methods. Furthermore, the high quality of the synthesized flames ($\textbf{CLIP Confidence = 0.76}$) establishes the Perlin mask as the most effective approach for wildfire image synthesis.

\subsection{Ablation Study}

\subsubsection{Style control}
It has been shown that the style of the final image is closely tied to the initial latent tensor. A key contribution of this study is the use of real images to control the overall style of the generated images. To validate this approach, we conducted experiments by generating images with and without real images as a style factor.

When real images were excluded, the generated masks exhibited poor quality. Even with higher noise ratios, which introduced more randomness, the resulting images appeared grayish and lacked the realism of natural scenes. The resulting images were far from resembling true wildfire scenes.

\begin{figure}[ht]
    \centering
    \centerline{\includegraphics[width=1\linewidth]{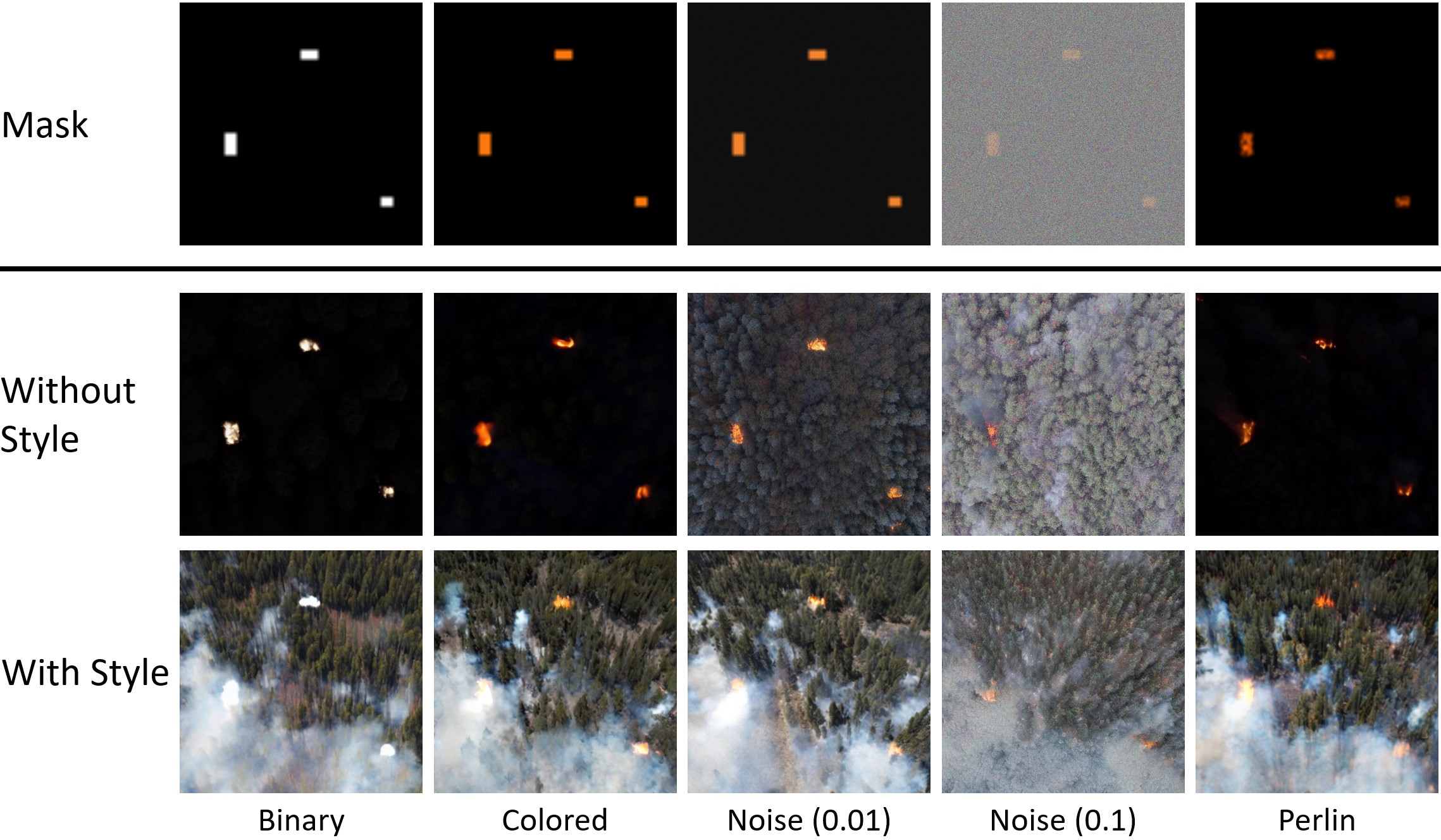}}
    \caption{Wildfire image synthesis with and without style image.}
    \label{fig:ablation}
\end{figure}

In contrast, when real images were used as part of the input, the generated images became much more realistic and maintained a consistent style with the original real image, even with noise injection, as shown in Figure \ref{fig:ablation}. 
Table \ref{tab:abl} further demonstrates the impact of incorporating real images as a style factor, with significant improvements in both image quality and flame realism when a real image is included in the input.

\subsubsection{Impact of noise}

The key purpose of the noise mask is to introduce additional variation during the denoising process. As shown in Figure \ref{fig:ablation}, incorporating a small amount of noise can enhance the visual quality of the synthesized flames compared to masks without noise. However, this method also has certain limitations. While noise can improve flame diversity, it tends to reduce overall image quality. Additionally, as the noise ratio increases, the quality of the flames progressively degrades.

\begin{table}[htbp]  
    \centering      
    \caption{Ablation Study}
    \resizebox{1\linewidth}{!}{
    \begin{tabular}{llll} 
        \toprule      
        (\textbf{without} style image)&nFID $\downarrow$& CLIP Score $\uparrow$& CLIP Conf. $\uparrow$\\\midrule
  Baseline& 0.48& 30.22& NA\\
 Binary Mask& 0.98& 14.31&0.58\\
   Colored Mask&        0.84& 17.17& 
0.62\\
 Noise Mask (best)& 0.74& 22.34& 
0.38\\
 Perlin Mask& 0.74& 18.02& 0.65\\
 & & &\\\toprule
 (\textbf{with} style image)&  nFID $\downarrow$& CLIP Score $\uparrow$& CLIP Conf. $\uparrow$\\\midrule
 Baseline&  0.41 (\better{-0.07})& 31.93 (\better{+1.71})& NA\\
 Binary Mask& 0.37 (\better{-0.61})& 26.61 (\better{+12.30})&0.38 (\less{-0.20})\\
 Colored Mask&  0.33 (\better{-0.51})& 31.28 (\better{+14.11})& 0.60 (\less{-0.02})\\
 Noise Mask (best)& 0.48 (\better{-0.26})& 31.87 (\better{+9.53})& 0.67 (\better{+0.29})\\
 Perlin Mask& 0.28 (\better{-0.46})& 31.53 (\better{+13.51})& 0.76 (\better{+0.11})\\\bottomrule
    \end{tabular}}
    \label{tab:abl}             
\end{table}

\subsubsection{Impact of domain warping noise (Perlin)}

The use of Perlin noise in this study is surprisingly suited for flame generation due to its smooth transitions and complex textures. Although the visual contrast between Perlin masks and simple colored masks may not be immediately apparent, the difference in actual image quality is substantial. As shown in Table \ref{tab:fid}, the Perlin mask method significantly outperforms the colored mask method in both overall image quality and flame realism. The seamless blending of Perlin noise with the background creates more realistic images, and the natural textures produced by Perlin noise lead to more lifelike flames, as mentioned in Section \ref{sec:noise}.

\section{Conclusion}
\label{sec: Conclusion}

In this paper, we introduced the \textbf{FLAME Diffuser}, a novel framework for wildfire image synthesis using mask-guided diffusion. The proposed method provides precise control over the placement of fire elements, allowing for the generation of realistic images that maintain the original style and context. By using real images and Perlin noise, our approach enhances both the quality and diversity of the generated wildfire images. The results, evaluated using nFID, CLIP Score, and CLIP Confidence, demonstrate that the FLAME Diffuser can produce high-quality images suitable for downstream tasks such as wildfire detection and monitoring. Our method's training-free nature and ability to synthesize large amounts of labeled data make it an effective tool for addressing data scarcity in this domain. Future work could explore the extension of this framework to other natural disaster scenarios and further refine the control over specific visual elements.


\section*{Acknowledgments}
\label{sec:Acknowledgement}

This material is based upon work supported by the National Science Foundation under Grant Numbers CNS-2204445.

\bibliography{references} 
\bibliographystyle{IEEEtran} 

\end{document}